# Phase-Only Planar Antenna Array Synthesis with Fuzzy Genetic Algorithms

Boufeldja Kadri [1], Miloud Boussahla [2], Fethi Tarik Bendimerad [2]

[1] Bechar University, Electronic Institute
P.O.Box 417, 08000, Bechar, Algeria

[2] Abou-Bakr Belkaid University, Engineering Sciences Faculty, Telecommunications Laboratory
P.O.Box 230, Tlemcen, Algeria

**Abstract**
This paper describes a new method for the synthesis of planar antenna arrays using fuzzy genetic algorithms (FGAs) by optimizing phase excitation coefficients to best meet a desired radiation pattern. We present the application of a rigorous optimization technique based on fuzzy genetic algorithms (FGAs), the optimizing algorithm is obtained by adjusting control parameters of a standard version of genetic algorithm (SGAs) using a fuzzy controller (FLC) depending on the best individual fitness and the population diversity measurements (PDM).
The presented optimization algorithms were previously checked on specific mathematical test function and show their superior capabilities with respect to the standard version (SGAs).
A planar array with rectangular cells using a probe feed is considered. Included example using FGA demonstrates the good agreement between the desired and calculated radiation patterns than those obtained by a SGA.

## 1. Introduction

Planar antenna arrays are fundamental components of radar and wireless communication systems [1]. Their performance heavily influences the overall system's efficiency and suitable design methods are necessary.

The phase-only methods are of particular interest in antenna array synthesis as phase shifters are used to control the direction of the main beam. These methods include in general nonlinear optimization algorithms.

The genetic algorithms (GAs) have been widely used in electromagnetic problems optimization, and particularly for the synthesis of antenna arrays. They have proved to be a useful and powerful alternative to traditional optimization techniques [2-7] when handling with multidimensional, multimodal optimization problems and their success are related to their versatility, robustness and their ability to optimize non differentiable cost function [2-7].

However, GA has also some demerits, such us poor local searching, premature converging as well as slow convergence speed. Adaptive genetic algorithms (AGAs) have been developed to overcome these problems, where their control parameters are adjusted according to the variation of the environment in which the GAs are run. We introduce the well-known performances of the fuzzy set theory to adjust control parameters of GAs depending on current performance measures of GAs such us: maximum, average, minimum fitness and on the diversity of the population(PD).

We present in this paper the synthesis of the complex radiation pattern of a planar antenna array with probe feed by only optimizing the phase excitation coefficients, the desired radiation pattern is specified by a narrow beam pattern with a beam width of 8 degrees and a maximum side lobe levels of -20DB pointed at 10°.

Section 2 describes the fuzzy genetic algorithms (FGAs), the design of a fuzzy controller is discussed to adjust crossover and mutation probabilities according to the population diversity measurements and the best fitness individual. Section 3 shows the synthesis problem of a planar antenna array with rectangular cells using FGAs by optimization of the phase excitation coefficients.

Numerical results for a planar array using both the SGAs and FGAs are presented in section 4, to compare the performances obtained while introducing fuzzy techniques in GAs. Finally, some conclusions are drawn in section 5.

## 2. Fuzzy Genetic Algorithms

The GAs behavior is determined by the exploitation and exploration relationship kept throughout the GA run. This balance between the utilization of the whole solution space and the detailed searching of some parts can be adapted to change of GA operators setting (selection, crossover and mutation). So, different genetic operators or control parameters values maybe necessary during the course of a run for inducing an optimal exploration/exploitation balance. For these reasons, adaptive GAs have been built that dynamically adjust selected control parameters or genetic operators during the course of evolving a solution [8] [9].

One way for designing AGAs involves the application of fuzzy logic controller (FLCs) [10-12] for adjusting GA control parameters.

The main idea of adaptive GAs based on fuzzy controllers FLCs is to use a FLC whose inputs are any combination of GA performances measures or current control parameters and whose outputs are GA control parameters. Current performance measures of the GA are sent to the FLC, which computes the new control parameters values that will used by the GA as demonstrated by the flowchart shown in figure 1.

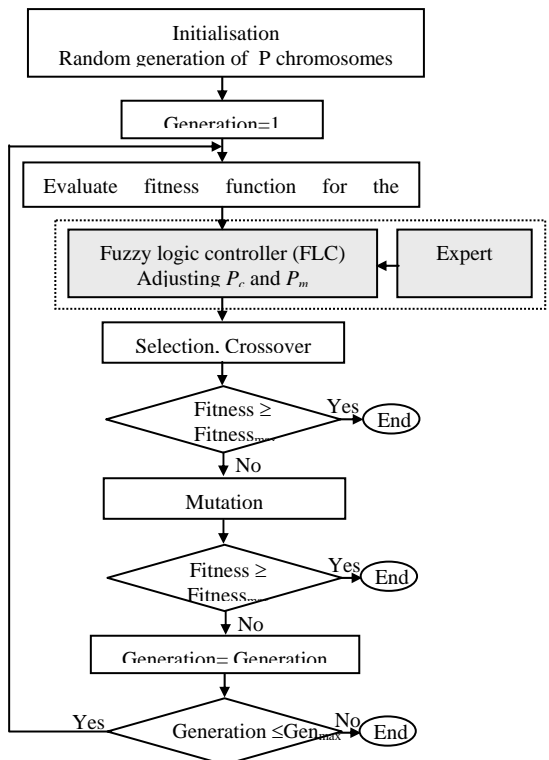

Fig. 1 Flowchart of the fuzzy genetic algorithms (FGAs).

FLC's inputs should be robust measures that describe GA behavior and the effects of genetic setting parameters and genetic operators, some possible inputs were cited in [9][10]: diversity measures, maximum, average, minimum fitness.

FLC's outputs indicate the values of control parameters or changes in these parameters, the following outputs were reported in [9] [10]: mutation probability ($p_m$), crossover probability ($p_c$), population size … etc.

We have choose for FLC's outputs the probabilities of crossover pc and mutation pm to realize the twin goals of maintaining diversity in population and sustaining the convergence capacity of the GA[13] [14].

The significance of $p_c$ and $p_m$ in controlling GA performance has long been acknowledged in GA research [6] [7]. Several studies, both empirical [15] [16] and theoretical [17] have been devoted to identify optimal parameter settings for GAs. The crossover probability $p_c$ controls the rate at which solutions are subjected to crossover. The higher the value of $p_c$, the quicker are the new solutions introduced into the population. As $p_c$ increases, however, solutions can be disrupted faster than selection can exploit them.

Mutation is only a secondary operator to restore genetic material choice. Nevertheless the choice of $p_m$ is critical to GA performance and has been emphasized in Dejong's work [18]. Large value of pm transforms GA into a purely random search algorithm, while some mutation is required to prevent the premature convergence of the GA to suboptimal solutions.

The FLC design takes into account the PDM and a performance measure of GAs, in this paper the FLC has three inputs ($D_{gw}$, $\overline{f}/f_{max}$ and Number) and two outputs (pc and pm) as indicated in the figure 2.

Where:

$\overline{f}$ : is the average fitness of the current population.

$f_{max}$ : is the fitness of the optimal individual.

$D_{gw}$ : is the gene inner diversity.

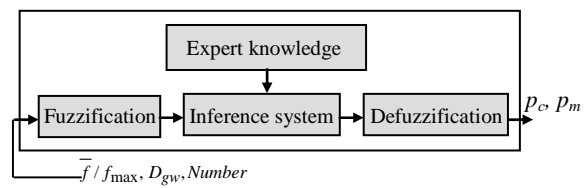

Fig. 2 Structure of the fuzzy logic controller FLC.

Let us consider a given population with M individuals ($p_1$ ,…, $p_M$) where each individual is represented by a binary string of $l$ bits, the PDM can be described by means of the gene inner diversity given by equation 1:

$$D_{gw} = \overline{\delta}_1 = \frac{1}{M.l} \sum_{i=1}^{M} \sum_{j=1}^{l} \left( p_i^j - \overline{g}^j \right)^2 \quad (1)$$

Where

$p_i^j$ : represents the j$^{th}$ bit gene value of the i$^{th}$ individual string.

$\overline{g}^j$ : is the gene average calculated by equation 2.

$$\overline{g}^j = \frac{1}{M} \cdot \sum_{i=1}^{M} p_i^j \quad (2)$$

$D_{gw}$ represents the genetic drift degree and evolution ability of current population. $\overline{f}/f_{max}$ is used to judge whether the current PD is useful [12], if it's near to 1, convergence has been reached, whereas if it's near to 0, the population shows a high level of diversity[10]. Number is used to record the frequency of the largest fitness value that is not changed.

The input variables $D_{gw}$, $\overline{f}/f_{max}$ and Number to be included respectively in the ranges : [0 , 0.25], [0 , 1] and [0 , 30].

Once the inputs and outputs of the FLC are defined, we must drive the membership functions and the fuzzy rules. More details about the design of FLC are given in [12].

## 3. Synthesis of Planar Antenna Arrays

We develop in this paper a synthesis of planar antenna array with probe feed using the FGAs discussed in the previous section.

Let us consider a planar antenna array constituted of MxN equally spaced rectangular antenna arranged in a regular rectangular array in the x-y plane, with an inter-element spacing of $d = dx = dy = \lambda/2$ as indicated by figure 3, and whose outputs are added together to provided a single output. Mathematically, the normalized array far-field pattern is given by:

$$F_s(\theta,\phi) = \frac{f(\theta,\phi)}{F_{s\max}} \cdot \sum_{m=1}^{M} \sum_{n=1}^{N} I_{mn} \cdot e^{j.(m-1)k_0.\sin\theta.\cos\phi.dx + j.\psi_{mn}}$$
$$\cdot e^{j.(n-1)k_0.\sin\theta.\sin\phi.dy} \quad (3)$$

Where
$f(\theta,\phi)$: Represents the radiation pattern of an element.

$I_{mn}$ : Amplitude coefficient at element *(m, n)*.

$\psi_{mn}$ : Phase coefficient at element *(m, n)*.

$k_0$ : Wave number.

If we consider an array with separable distribution, then the array factor is the product of two linear arrays associated with the row and column direction of this planar, which can be expressed in the form (4):

$$F_s(\theta,\phi) = \frac{f(\theta,\phi)}{F_{s\max}} \cdot \sum_{m=1}^{M} I_m \cdot e^{j.(m-1)k_0.\sin\theta.\cos\phi.dx + j.\psi_{mn}}$$
$$\cdot \sum_{n=1}^{N} I_n \cdot e^{j.(n-1)k_0.\sin\theta.\sin\phi.dy} \quad (4)$$

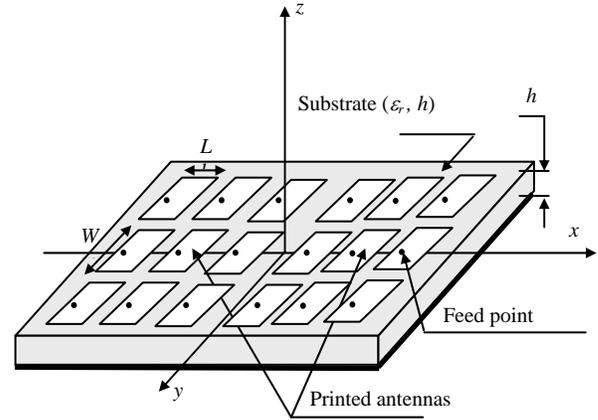

Fig. 3 Planar antennas array fed by coax.

We use the FGAs to find the complex excitation coefficient vector

$$A = \left[\psi_{x1}, \psi_{x2}, ..., \psi_{x\frac{M}{2}}, \psi_{y1}, \psi_{y2}, ..., \psi_{y\frac{N}{2}}\right]$$

so the radiation pattern produced satisfy the desired radiation pattern specified by the pattern model as illustrated in figure 4. This pattern has a narrow beam with –20DB sidelobes. The pattern is normalized to the peak value at 10 degrees and must have a 3DB beamwidth of at least 8 degrees. The –20DB sidelobe level must be met beginning at 0 and 20 degrees and extending to ±90 degrees. The sidelobes in this case are defined relative to the peak of beam at 10 degrees. The specifications are illustrated in figure 4.

We have choose a suitable fitness function that can guide the SGAs and FGAs optimization toward a solution that meets the desired radiation pattern as mentioned in [1]. Equations 5-7 describe the appropriate fitness function.

$$d_{av} = \frac{1}{2S+1} \sum_{i=-S}^{S} d_i \quad (5)$$

$$Sll_{\max} = \min_{\forall i \in Sidelobes} \left(|d_{av} - d_i|\right) \quad (6)$$

$$fitness = d_{av} + w_1 Sll_{\max} \quad (7)$$

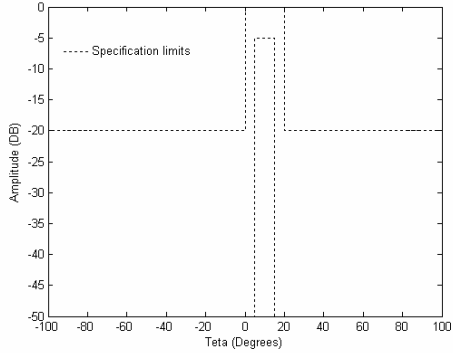

Fig. 4 Plot of the desired pattern specification (for 10° main beam).

Where it is assumed that a number of samples of the pattern, $d_i$ in $dB$, are taken in the beam region and the sidelobe region and that the number of samples in the beam region is equal to $2S+1$. An arbitrary weight $w_l$ is used. The goal of function (7) is to maximize the difference between the average value in the beam and the highest sidelobe.

First we have to find a relationship between the GAs and the array. In the case of a coded GA, each element of the array is represented by a string of bits which gives the complex excitation of the element; hence each element is characterized by its phase excitations. This relationship is shown in table 1.

Table 1: The relationship between elements of GAs and arrays.

| Genetic parameters | Antennas array |
|---|---|
| Gene | Bits chain(string): (phase) |
| Chromosome | One element of array |
| Individual | One array |
| Population | Several arrays |

## 4. Numerical Results

In our simulation, we have used a population size of 40 for GAs. Roulette strategy for "selection" one–point crossover and mutation to flip bits. For the SGAs, we have used value of $p_c$=0.71 and $p_m$=0.02, for the FGAs, $p_c$ and $p_m$ are determined according to FLC presented previously.

We have chosen for simplification a symmetrical array, whose elements are located symmetrically on x-y plan, and adopted an antisymetrical phases for elements, which can be resumed by equation (7):

$$\begin{cases} x_i = -x_{-i}, \ \psi_i = -\psi_{-i} \\ y_j = -y_{-j}, \ \psi_j = -\psi_{-j} \\ \text{for } i = 1,...N/2, \ \text{for } j = 1,...M/2 \end{cases} \quad (8)$$

We have adopted a desired radiation pattern specified by a narrow beam pointed at 10 degrees with a sidelobe level of -20dB. Figures 5 to 8 show the synthesis result of a probe-fed planar array constituted by 8x16 half wavelength spaced rectangular microstrip antennas with 0.906cm width and 1.186cm long working at the frequency of 10GHz.

In figure 5 we present the result of planar array optimization by phase excitation coefficients using both SGAs and FGAs. It is clearly seen that the radiation pattern obtained by FGAs meet better the desired pattern than the obtained by SGAs. The sidelobe level obtained by FGAs optimization (-26DB) are much better than in the case of SGAs (-20DB).

From figure 6, the speed approaching the global optimal of FGA is much quickly than that of SGA, and the fitness values of the best individuals of FGA are almost higher than that of SGA in every population. For each generation the probabilities $p_c$ and $p_m$ are adjusted according to the response of the fuzzy controller, and shown in the figures 7 and 8.

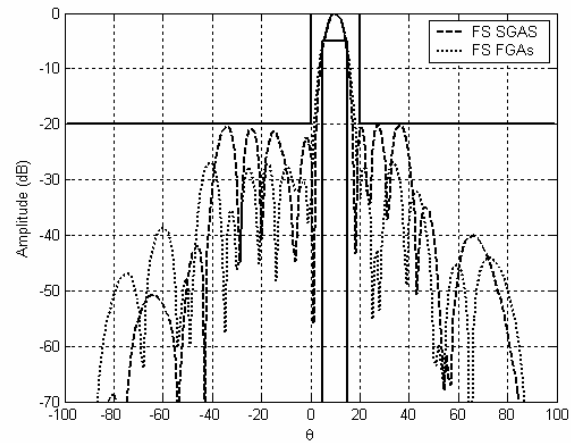

Fig. 5 Result of a planar array synthesis with 8x16 rectangular microstrip antennas applying both SGAs and FGAs.

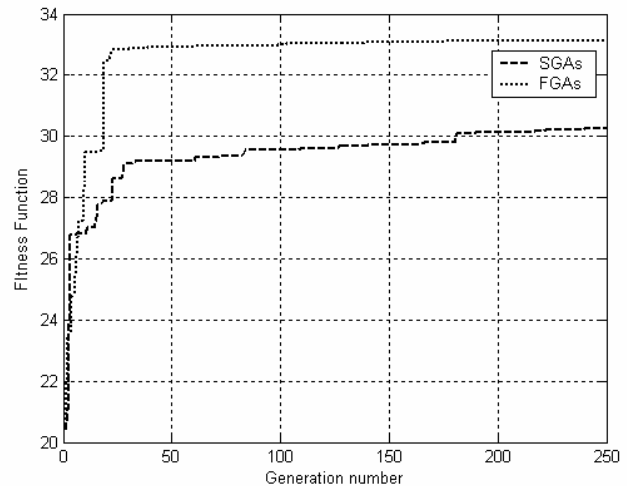

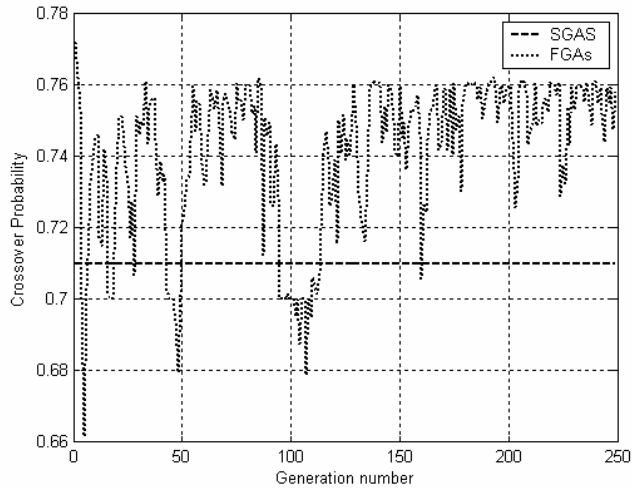

Fig. 6 Comparison between fitness functions obtained by the two algorithms SGAs and FGAs.

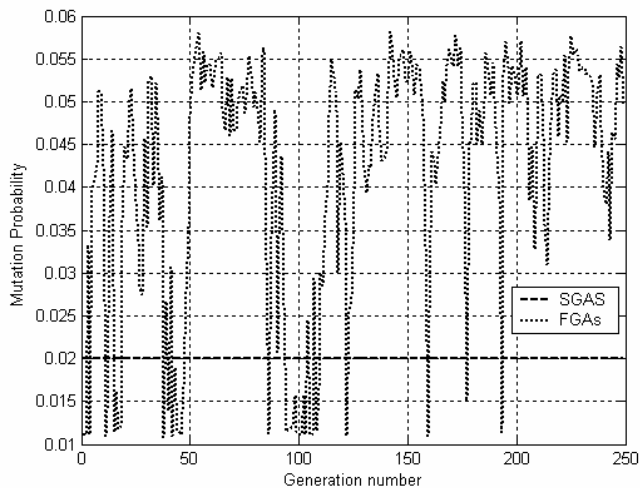

Fig. 7 Adjusting $p_c$ during GAs run.

Fig. 8 Adjusting $p_m$ during GAs run.

## 5. Conclusions

A rigorous method for the synthesis of planar antenna array using AGAs integrating a FLC by optimizing only phase excitation coefficients has been presented. The GAs behavior is strongly determined by the balance between exploiting what already works best and exploring possibilities that might eventually evolve into something even better.

The balance between these characteristics (exploration and exploitation) of the GAs is dictated by the values of pc and pm. We have adopted the variation of pc and pm according to the response obtained by a FLC which depends on the PDM and a measure of the convergence by means of the ratio between the best fitness and average fitness. With the approach of adaptive probabilities of crossover and mutation, we also provide a solution to the problem of choosing the optimal values of the probabilities of crossover and mutation for the GA.

From the simulating results, it has been shown that the speed approaching the global optimal of FGA is much quickly than that of SGA, and the fitness values of the best individuals of FGA are almost higher than that of SGA in every population.

## References

[1] R. L. Haupt, J. M. Johnson, "Dynamic Phase-Only Array Beam Control using a Genetic Algorithm",1st NASA/DOD Workshop on Evolvable Hardware 217-224, EH'99, July 19-21, Pasadena , CA, USA
[2] R. L. Haupt, "An Introduction to Genetic Algorithms for Electromagnetics", IEEE Antenna and propagation Magazine, Vol. 37, pp. 7-15, 1995.
[3] S. A. Mitilineos, C. A. Papagianni, G., I. Verikaki, C. N. Capsalis, "Design of Switched Beam Planar Arrays Using the Method of  Genetic Algorithms", Progress In Electromagnetics Research, PIER 46, 105-126, 2004.
[4] D. Marcano, F. Duran, "Synthesis of Antenna Arrays Using Genetic Algorithms", IEEE Antenna and propagation Magazine, Vol. 42, NO. 3, June 2000.
[5] M. Donelli, S. Coarsi F. De Natale, M. Pastorino, A. Massa, "Linear Antenna Synthesis with Hybrid Genetic Algorithm", Progress in Electromagnetics Research, PIER 49, 1-22, 2004.
[6] D. E. Goldberg, "Genetic Algorithms in Search, Optimization and Machine Learning", Reading, MA: Addison Wesley, 1989.
[7] K. A. Dejong, "Genetic Algorithms: A 10 year perspective", in Proceedings of an International Conference of Genetic Algorithms and Their Applications, (J Greffenstette, editor), Pittsburgh, July 24-26, 1985, PP. 169-177.
[8] M. Srinivas, L. M. Patnaik, "Adaptive Probabilities of Crossover and Mutation in Genetic Algorithms", IEEE Trans. Syst. Man and Cybernetics, 1994, 24(4): 656-667.
[9] F. Herrera, M. Lozano, "Adaptive Genetic Operators Based On Coevolution with Fuzzy Behaviors", IEEE Transaction on Evolutionary Computation, Vol. 5, NO. 2, 2001.
[10] M. A. Lee, H. Takagi, "Dynamic Control of Genetic Algorithms Using Fuzzy Logic Techniques", International Conference on Genetic algorithms ICGA'93, Urbana-Champaign, pp. 76-83,1993
[11] B. Kadri, F.T. Bendimerad, "Fuzzy Genetic Algorithms for The Synthesis of  Unequally Spaced Microstrip Antennas Arrays", European    Conference on Antennas And Propagation  EUCAP2006 , Nice, 6-10 November 2006, (ESA SP-626 , October 2006).
[12] K. Wang, "A New Fuzzy Genetic Algorithm Based on Population Diversity", International Symposium on Computational Intelligence in Robotics and Automation, pp. 108-112, July 29 August 1, 2001, Alberta, Canada.

[13] Xiagofeng Qi, Francesco Palmieri, "Theoretical Analysis of Evolutionary Algorithms with an Infinite Population Size in Continuous Space, Part II: Analysis of the Diversification Role of Crossover", IEEE Trans on Neural Networks, 1994, 5(1)

[14] Z. Liu, J. Zhou, Z. Wei, H. Lv, L. Tao, "A Study on Novel Genetic Algorithm with Sustaining Diversity", Proceeding of ICSP2000, 1650-1654.

[15] J. J. Greffenstette, "Optimization of Control Parameters for Genetic Algorithms", IEEE Trans. Syst. Man., and Cybernetics, Vol. SMC-16, No. 1, pp. 122-128, Jan/Feb, 1986.

[16] J. D. Schaffer et, al., "A Study of Control Parameters Affecting online Performance of Genetic Algorithms for Function Optimization", Proc. Third Int. Conf. Genetic Algorithms, 1989, pp. 51-60.

[17] J. Hesser, R. Manner, "Towards an Optimal Probability for Genetic Algorithms", Proceeding of the First Workshop, PPSN-I, pp. 23-32, 1990.

[18] K. A. Dejong, "An Analysis of the Behavior of a Class of Genetic Adaptative Systems", Ph.D. Dissertation, University of Michigan, 1975.

**Boufeldja Kadri** was born in Bechar, Algeria, in 1972. He received the Majister degree in 1998, from the Abou Bekrbelkaid University in Tlemcen (Algeria). Since 1999, he joined the Electronic Institute in Bechar University (Algeria), where he is now an associate professor. His research interests include modelling and optimization of antenna array, heuristic algorithms.

**Miloud Bousahla** was born in Sidi BelAbbès, Algeria, in1969. He received the Magistère diplomas in 1999 from Abou BekrBelkaıd University in Tlemcen (Algeria). He is currently a Junior Lecturer in the Abou BekrBelkaıd University. Also he is a Junior Researcher within the Telecommunications Laboratory. He works on design, analysis and synthesis of antenna and conformal antenna and their applications in communication and radar systems.

**Fethi Tarik Bendimerad** was born in Sidi BelAbbès, Algeria, in1959, he received his Phd degree from the Sophia-Antipolis University in Nice (France), in1989. He is currently a full professor at the Faculty of engineering at the Abou BekrBelkaıd University in Tlemcen, (Algeria) and the Director of the Telecommunications Laboratory. His field of interest is antenna treatment and smart antenna.